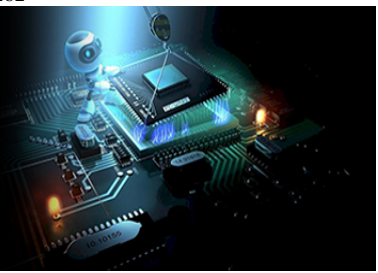

**Aya Kaysan Bahjat**
Informatics Institute for Postgraduate Studies, Baghdad, Iraq

# Text images processing system using artificial intelligence models

**Aya Kaysan Bahjat**

**DOI:** https://www.doi.org/10.33545/26633582.2025.v7.i2c.222

**Abstract**
This is to present a text image classifier device that identifies textual content in images and then categorizes each image into one of four predefined categories, including Invoice, Form, Letter, or Report. The device supports a gallery mode, in which users browse files on flash disks, hard disk drives, or microSD cards, and a live mode which renders feeds of cameras connected to it. Its design is specifically aimed at addressing pragmatic challenges, such as changing light, random orientation, curvature or partial coverage of text, low resolution, and slightly visible text.
The steps of the processing process are divided into four steps: image acquisition and preprocessing, textual elements detection with the help of DBNet++ (Differentiable Binarization Network Plus) model, BART (Bidirectional Auto-Regressive Transformers) model that classifies detected textual elements, and the presentation of the results through a user interface written in Python and PyQt5. All the stages are connected in such a way that they form a smooth workflow.
The system achieved a text recognition rate of about 94.62✻% when tested over ten hours on the mentioned Total-Text dataset, that includes high resolution images, created so as to represent a wide range of problematic conditions.
These experimental results support the effectiveness of the suggested methodology to practice, mixed-source text categorization, even in uncontrolled imaging conditions.

## 1. Introduction
Our paper is an attempt to present a text-image categorization system that plays a key function in the present-day information management in terms of organizing, indexing and routing huge heaps of written and scanned data, and therefore allows an organization to access and take action on information in a rapid and dependable manner. conventional manual document classification, which is normally done by clerical personnel/personnel or subject matter experts, inconsistently done, is time consuming, and may lead to error as well as prejudice, and thus undermines the decision making process and swells the operational expenses as the size of text image collections increases. In order to overcome these limitations, automated approaches utilize characteristics found in the text images and their outlay and visual representation for assigning labels or categorize them with high throughput and reproducibility.
In modern academic literature, the spectrum of the classification pipelines contrasts the traditional rule-based frameworks, and the classic statistical learning procedures, i.e. feature engineering, combined with Naive Bayes, support-vector machines, or logistic regression, with the latest paradigm of deep learning, where hierarchical and contextual representations are learned directly using raw data. The latter include convolutional neural networks, recurrent architectures and, more recently, encoder models based on transformers.
 It is clear that automated systems enhance scalability, reduce human effort, and enable delivery of services in a timely fashion, like intelligence search, automated routing, compliance checks and analytics, and simultaneously reduce labeling discrepancies and latency in retrieval. However, some of these obstacles exist, i.e., heterogeneous text, imbalance in classes, noise introduced by OCR, and the need of annotated training data [3].
The remaining manuscript has following structure: Section 2 is a literature review of topics in this field; Section 3 explains the models used architecture, and Section 4 reports the outcome of the experiment in text detection and analyzes the Total-Text database after it has

**Corresponding Author:**
**Aya Kaysan Bahjat**
Informatics Institute for Postgraduate Studies, Baghdad, Iraq

been adapted to the English language (see Fig. 1) [4]; and, finally, Sections 5 and 6 give certain conclusions about the model and recommend the early-career researchers.

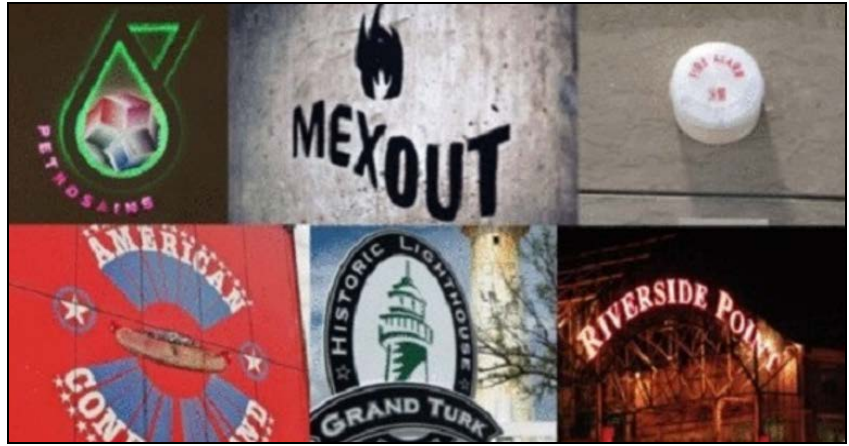

**Fig 1:** Total-Text Dataset Over View [4].

## 2. Related Work
S. Long, *et al*., in 2018 [5], The Text-Snake model is a model of text demonstration of scene which is seen as the configurations of disks which are overlapped and which are spread along the central axis of the text. It uses a fully convolutional network (FCN) to estimate geometric features including radius and orientation. The mechanism has been found to be better in the simulation of curved text with a set of benchmark models on the Total-Text, SCUT-CTW1500, ICDAR2015, and MSRA-TD500 datasets.

Text Snake produces an F -score change of more than 40 per cent in comparison to the baseline on Total -Text, an important result highlighting its effectiveness in identifying the forms of free-text. Y. Baek, *et al*., in 2019 [6] CRAFT Character Region Awareness to Text model is a character-based heatmap with affinity mapping to show the textual boundaries between a wide range of texts through complex segmentation methods. When applied in an empirical study on its F-measure on the Total-Text dataset, its empirical performance is approximately 83.6 percent, which highlights its strength in handling very irregular text structures.

It is based on this that later studies have built up on this to refine both of the pre-processing methods, like blind deconvolution, and postprocessing methods, like automatic blur classification, for further improve the detection in a variety of natural scene configurations, J. Ye, *et al*., in 2020 [7] This system follows an instance-segmentation detection system with synthesis of features at character, word and global levels. With a single scale at which a ResNet on a backbone can run, it has an F-measure of 87.1 on the Total-text corpus, S. Zhang, *et al*., in 2021 [8] They used Adaptive Boundary Proposal Network of Arbitrary Shape Text Detection (ABPNet) that overcomes the problem of detecting arbitrary shape text using systemic method of generating a proposal boundary and externalizing it via iterative boundary deformation mechanism. In this context, a lightweight proposal boundary model takes rough edges of the text, and an adaptive deformation network, trained through Graph Convolutional Network (GCN) or Recurrent Neural Network (RNN) blocks, distorts the rough edges to perfectly fit the shape of the text behind them. Therefore, the architecture eliminates the overuse of elaborate post-processing effects, and increases the accuracy of boundaries around highly curved words. The ABPNet performance on the Total-Text dataset is provably high with high precision and recall values resulting in an F -measure is 86.87; such an outcome therefore shows learned boundary deformation as a viable alternative to dense masking or polygonic regression paradigms, F. Zhao, *et al*., in 2022 [9] The approach drives on a ResNet -50 backbone whose architecture is synergistically coupled with global and world-level cues in the detection of scene texts. Test run on the Total-Text corpus recorded an impressive F -score of 87.9 percent that outperforms the previously best available detectors under similar settings, Y. Su, *et al*., in 2022 [10] Discrete Cosine Transform (DCT). The Discrete Cosine Transform (DCT) is a more theoretically parsimonious model of text instance detection, which projects discrete masks onto small-scale vectors, using the DCT. The resulting approach achieves a throughput of 15.1 frames per second and an F -score of 84.9 percent on the Total-Text benchmark hence provides a good trade off between accuracy and computing power, Y. You, *et al*., in 2023 [11] It is based on the idea of deformable Detection Transformer (DETR) that is required for directly estimate control B-Spline coordinates of every textual contour, thus permitting smooth and highly flexible modeling of arbitrarily-shaped lexical units. Analysis of the Total Text corpus indicates that an F -measure of 87.6 per cent is attained without prior training data, hence demonstrating strong contour accuracy at increased parametric complexity and computational cost, Z. Chen, *et al*., in 2024 [12] Such an approach builds on a deformable Detection Transformer (DETR) architecture, which manipulates B -Spline control points directly on text contours, thus allowing flexibly and smoothly modeling words with arbitrary shapes. On the Total-Text benchmark it scores an F -measure of 87.6, highlighting the better contour accuracy without using pre-training data, but with the cost of more complex parameters and higher computation cost.

We have observed that the high accuracy expectations were often not reached in the past even when the studies were conducted in non-constrained situations which included things like different illumination, different orientations, curved texts, partial occlusions, low-resolution images and as well as distant viewpoints. We therefore tested our system on the Total-Text data which replicates such tough conditions. The major paper objectives is to achieve high precision in both real-time and browsing mode of text detection.

**3. Methodology:** In this section, our project details are illustrated, and it's structure shown in Fig. 2,

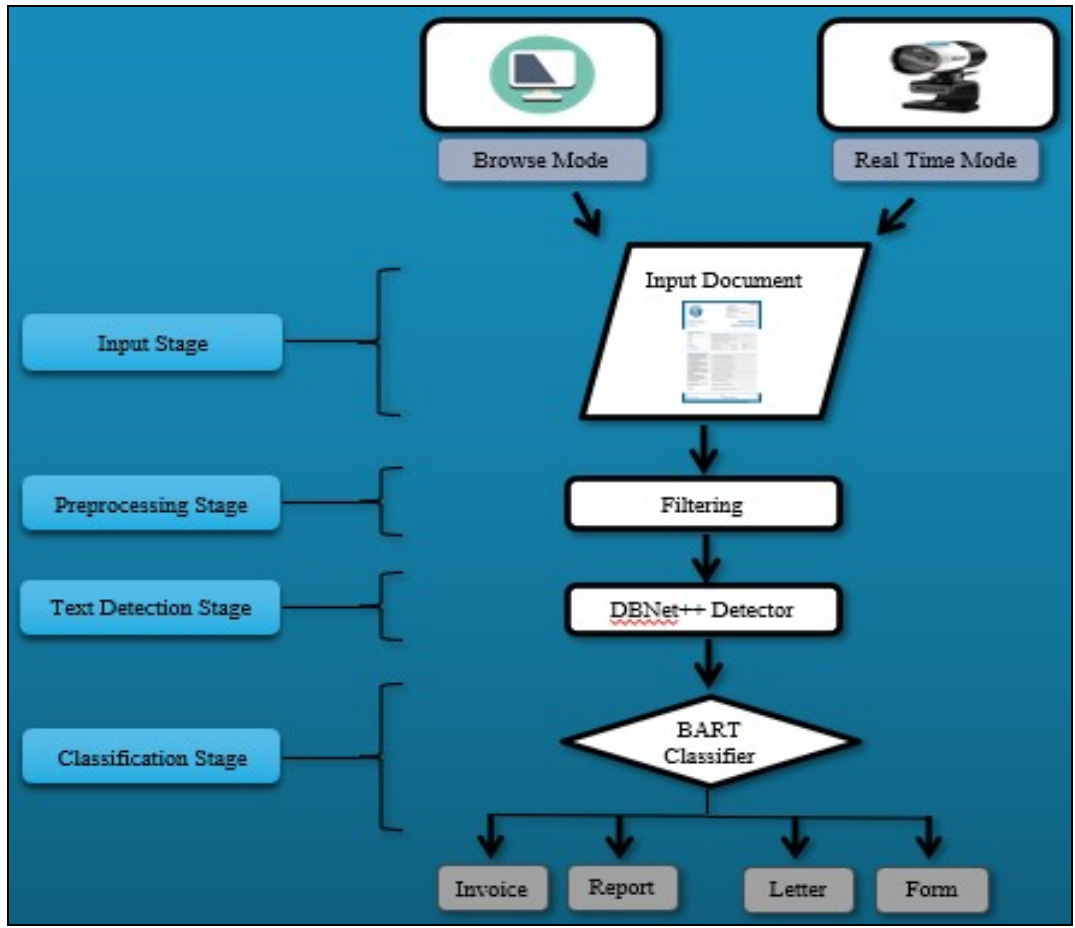

**Fig 2:** Automated Classification System Block Diagram.

In our system we utilize four entity classes that are predefined and include Invoice, Report, Letter, and Form to classify input data. First, the photographs are converted into grayscale (as cited in reference 13), and then are improved with the help of RealESRGAN (reference 14) and Contrast Limited Adaptive Histogram Equalization (CLAHE) (reference 15). Then, text segmentation is obtained through Differentiable Binarization Network Plus (DBNet++) that is an expansion of a ResNet-50 framework along with a differentiable binarization (DB) submodule. It is an architecture that is jointly optimised on a text probability map and a threshold map within a completely end-to-end training regime. Lastly, the classification of a text is done by using a Bidirectional Autoregressive Transformer (BART) [17], a model that uses encoder and decoder networks, and is further designed with Softmax layer. When the proper diligence is applied in the use of these models, the tasks such as text classification and summarisation can be easily done as shown in Figure 3. The text is, therefore, categorized into one of the above mentioned categories based on the results of classification step.

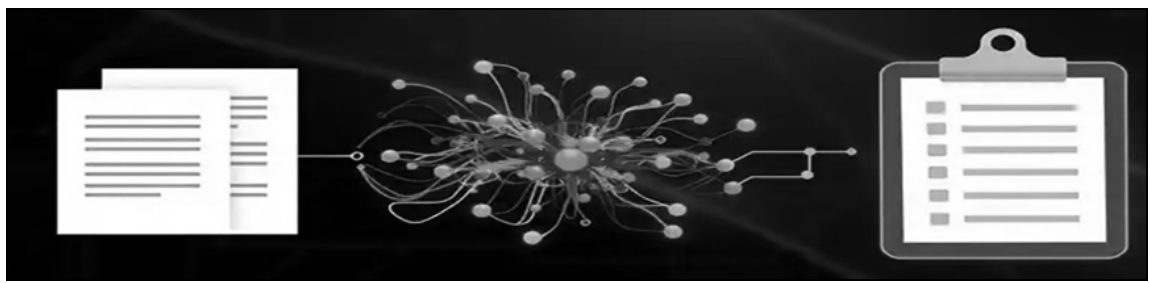

**Fig 3:** BART Rule over View [17].

**3.1 Preprocessing Operations**
By utilizing RealESRGAN, and CLAHE filters, we can preprocess the input image as follow:

**1. RealESRGAN**
In Fig. 4 after convert input image into grayscale, RealESRGAN [14] will increase its resolution.

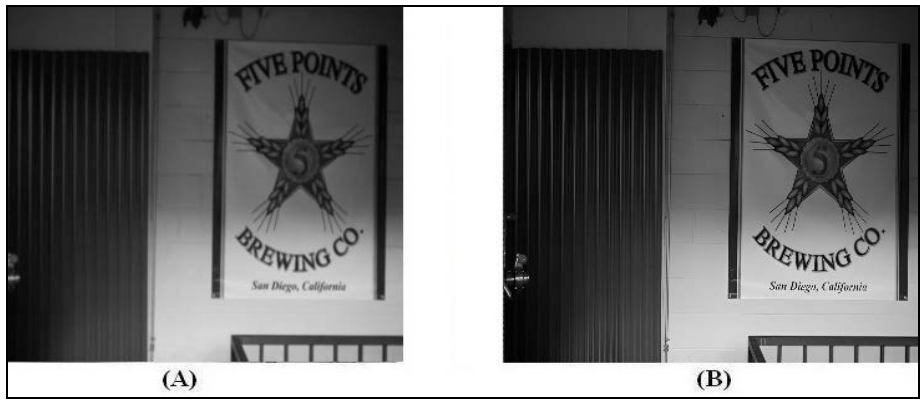

**Fig 4:** (A) Original Image (B) Image with Applying RealESRGAN Filter.

In this paper, a super-resolution image generator known as RealESRGAN_x2 has been used for increasing spatial resolution of the original raster via twice, which increases the legibility of fine strokes, narrow serifs, and other character elements that have a low-contrast threshold. This up-scaling operation enhances sturdiness of further optical character recognition (OCR) algorithms as shown in the Proposed Method section. A more pragmatic version, known as Real -ESRGAN, uses a Residual in Residual Dense Block (RRDB) generator and is trained on an abstract degradation pipeline. The model therefore generalises well to the kind of artefacts that are usually experienced in real world imagery, which include compression artefacts, sensor noise and blur. RRDB architecture with degradation-aware training guarantees stroke continuity and contrast preservation and limits the amount of unnecessary hallucinated texture introduced to the image, which may cause confusion in the context of the OCR processing.

In the proposed system, we utilize the RealESRGAN_x2 as the type of text detectors that is upscaling; the parameters that we use are: scaling factor of two, using the RealESRGAN_x2 model, execution in real-time, tiled inference, a tile dimension of 64, and tile overlap of 16, which meets the memory and latency requirements.

The parameters have been selected to balance the quality of restoration with the cost of computation but were biased in conserving the nature so that the character legibility is improved as much as possible without introducing artificial detail.

**2. Contrast-Limited Adaptive Histogram Equalization (CLAHE):** The next stage would involve the Contrast Limited Adaptive Histogram Equalization (CLAHE) application [15] filter whereby an image-specific clip factor is used to avoid exaggeration of noise in homogenous areas in the histogram; whereby any dominant peak would be clipped based on a pre-established threshold. This paper has used a clip limit of 8.0 and 8x8 tile size. The contrast enhancement is aimed at making textual features more visible (as seen in Figure 5), particularly when the light is low, and therefore, it is expected to make the algorithm more effective in capturing them.

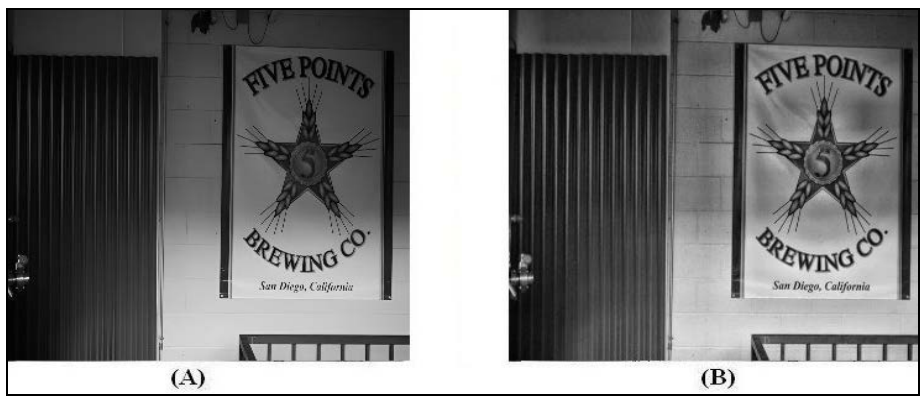

**Fig 5:** (A) Original Image (B) Image with Applying CLAHE Filter.

**3.2 DBNet++ Detector**
The state-of-the-art in text detection segmentation is used in the study, namely DBNet++ [16], due to its ability to localize arbitrarily-shaped textual elements. The architecture employs a backbone which is the ResNet-50 with a feature pyramid network (FPN) thus producing multi-scale feature maps that provide detailed representations of the input imagery. Then, these feature maps are propagated by a differentiable binarization module which learns in an end-to-end manner a text probability map and an adaptive threshold map. Applicant text areas are categorised by using a learnt soft threshold: pixels with probability higher than the identified threshold value are grouped together into connected components, thus processing not the entire image at once but a single text instance. As per the stipulations of this research, minimum and maximum text heights of 5 and 1024 pixels respectively were enforced so as to make it robust in scales and the default threshold value of 0.25 was picked to use in the preliminary candidate selection.

Three salient criteria were used to decide on the choice of DBNet++. First, it includes a differentiable binarization process, thus guaranteeing the accurate outlining of both curved and irregular textualizations. Secondly, the model has a strong ability to detect the character on a wide range of scales because of its multi-scale character of feature aggregation. Third, in addition to demonstrating high accuracy of detection, it also provides rich inferences of features that can be deployed in real-time.

In the framework of this investigation, DBNet++ is implemented in the PyTorch ecosystem with the MMOCR toolbox being used to make the integration easier. Figure 6 illustrates a typical example of text recognition with the usage of DBNet++.

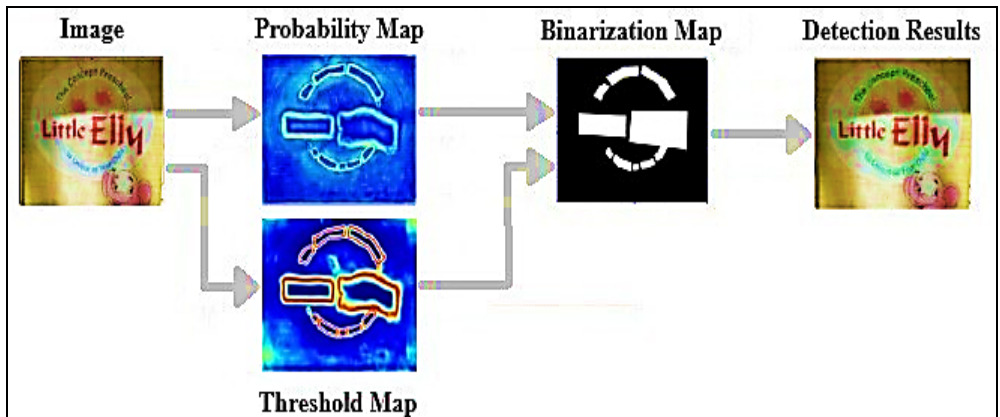

**Fig 6:** DBNet++ Detector.

**3.3 BART Classifier:** As it was previously stated, the Bidirectional and Auto-regressive Transformers (BART) structure is used to carry out text classification, taking advantage of its encoder-decoder paradigm. Specifically, one decided to use the facebook/bart -large -mnli variant because it is optimised towards natural language inference (NLI), thus allowing powerful zero-shot classification. In the inference stage, the encoder takes the input sequence, be it a linguistic description of an image or a bare sentence and encodes it into a sequence of contextualised token embeddings, which are successively optimised by stacking many layers of self-attention. Later, the embeddings obtained in the previous layer are fed to a pre-trained decoder that is trained with a natural language inference (NLI) paradigm by the systematic generation of hypotheses, and the text at hand, namely, the label [label], is refined. The decoder uses cross-attention across its sequential layers to reconstruct the representations of generated tokens with the original encoder representations, thus yielding an accurate mapping of the original encoder representations. The end of this architecture is a softmax layer, which, given each hypothesis of the output, compares the probability distribution under three canonical NLI labels: entailment, neutral, and contradiction. The ultimate classification decision is therefore obtained by choosing the label which achieves the highest probability which is entailment. This methodological aspect provides BART model with ability to make correct prediction of the classes without any task specific supervision, making it well appropriate in zero-shot classification problems as demonstrated in Figure 7.

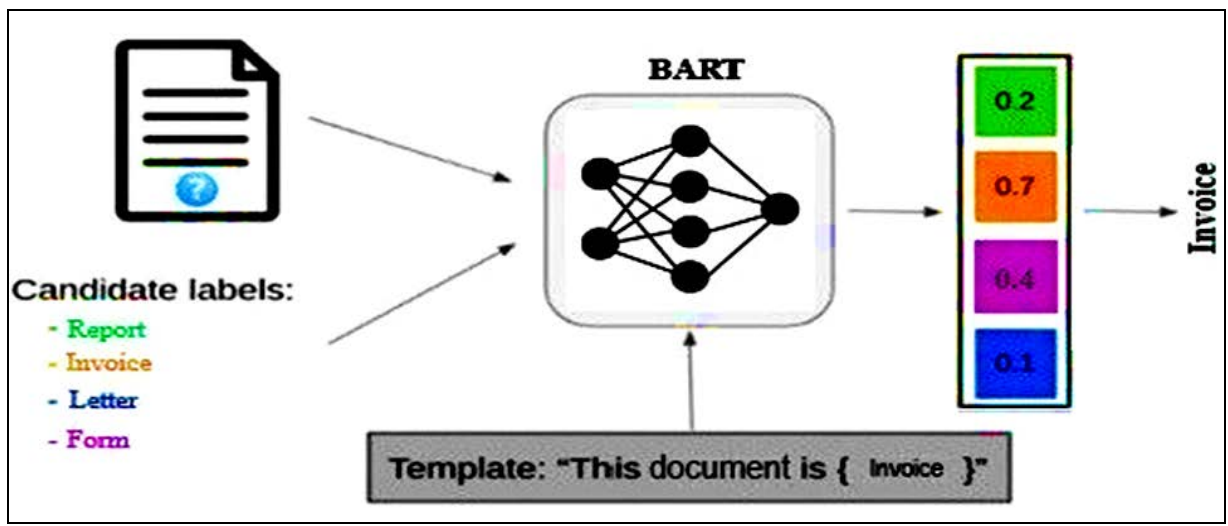

**Fig 7:** Time and Success Ratio evaluation.

**4. Experimental Result**
In the proposed project, the pretrained models were called DBNet++ which was utilized for detecting text in images, while BART variants (facebook/bart-large-mnli to classify and facebook/bart-large-cnn to summarize) were used, thus eliminating the need to undergo a training processing stage. Conversely, the testing stage took a total of about ten hours to identify all the text instances using the complete set of images in the Total-Text dataset and a text detection rate of 92.88 percent was obtained as shown by Figure 8.

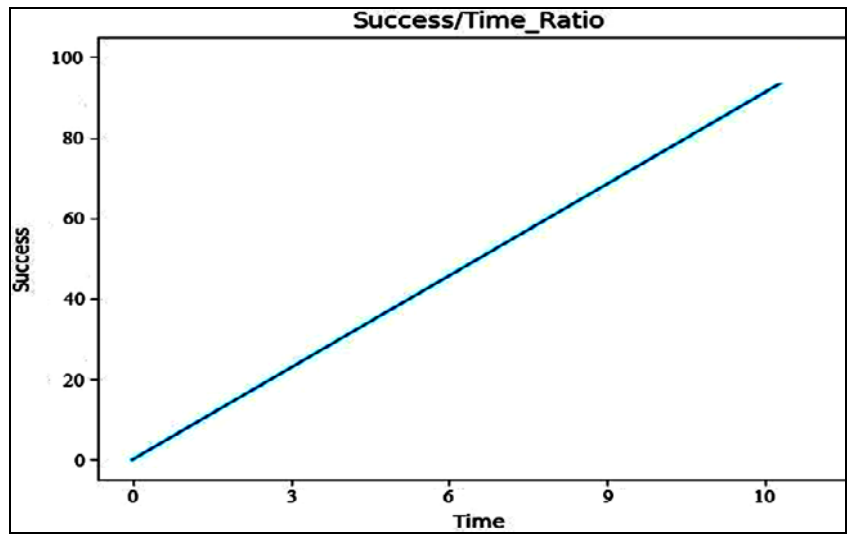

**Fig 8:** Time and Success Ratio evaluation.

The use of the built-in camera of the laptop or a webcam is possible. In this we have chosen the Microsoft Lifecam Studio (see Fig. 9), and this has made it possible to perform real-time experiments.

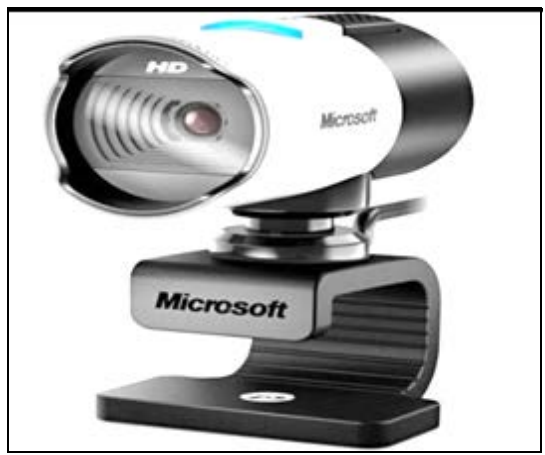

We give an experimental example in browse mode, Figure 10, of the process of text detection and classification, and one in real-time mode, Figure 11; the two are of course different but related in our overall approach.

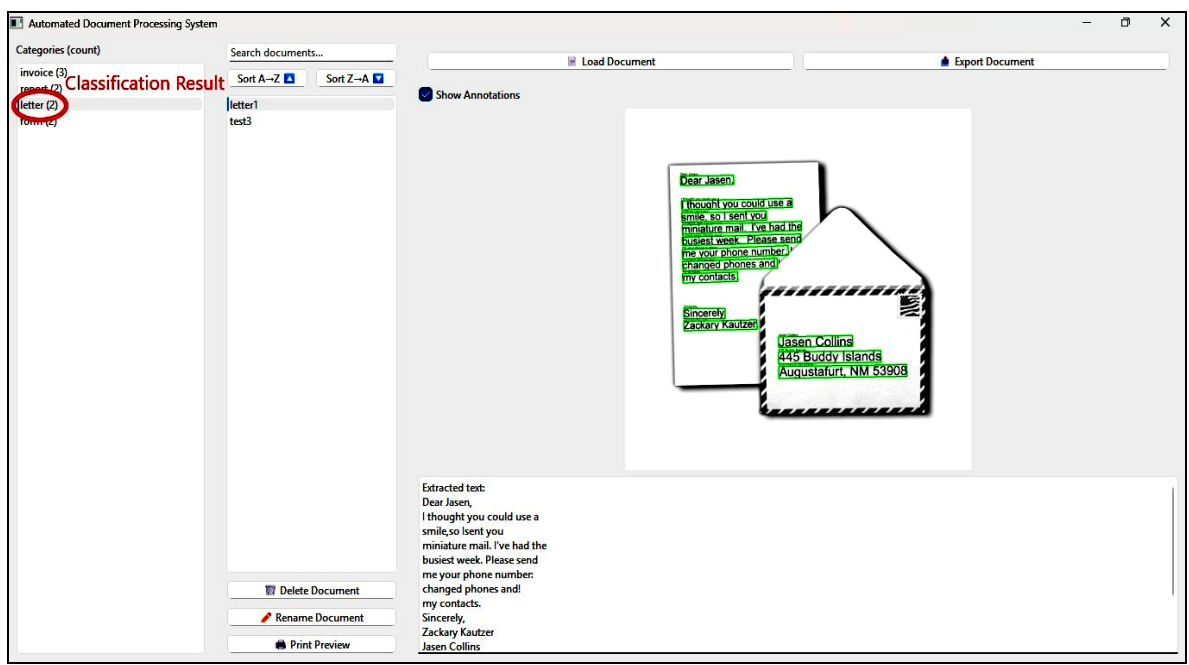

**Fig 10:** Browsing Mode Example: Classified Text in Input Image

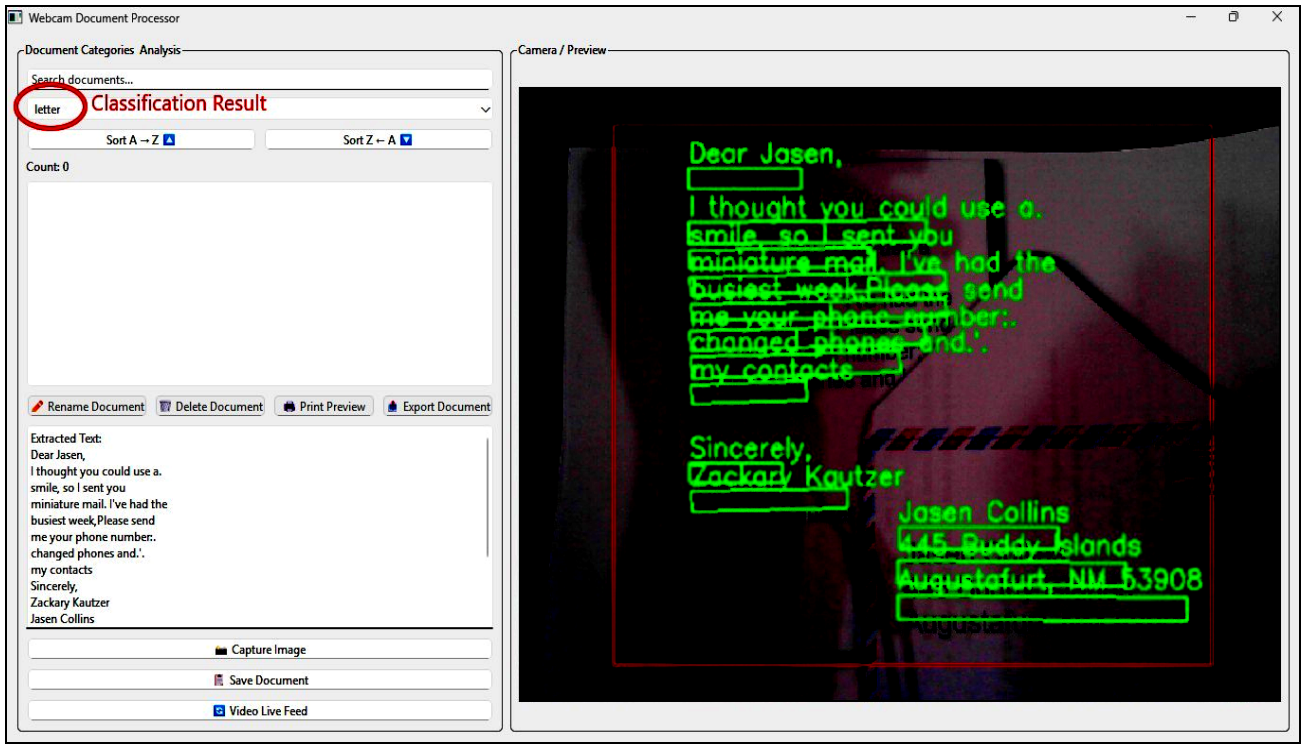

**Fig 11:** Real-time Mode Example: Classified text in Input Frame.

## 5. OCR Advantages [18]
Textual content of images is detected and this aids organizations in various fields (see Figure 12). This ability improves the working processes, increases the productivity, and supports the compliance with the regulations. The text-image-detection is applied in industry and the following case studies can be used to explain the application:

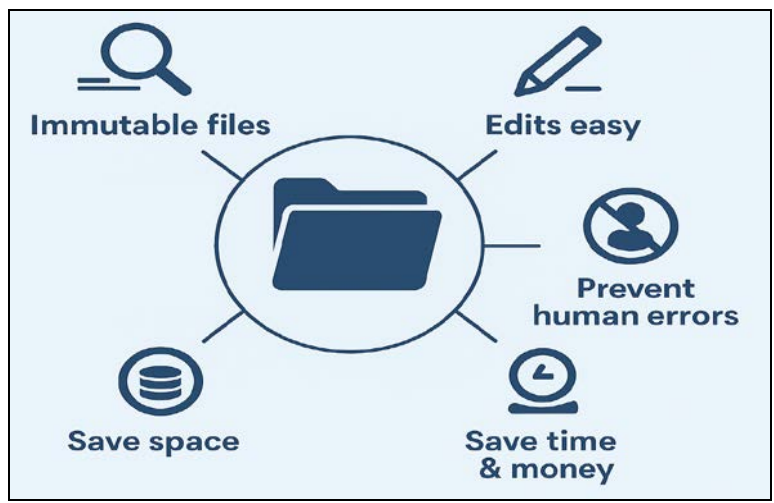

**Fig 12:** Texts Images Detection in Real-Life Use Cases [17].

**Make Immutable Documents more searchable.**
In cases where organisations have large libraries of PDF documents and textual electronic images, the data stored in the records is often non-searchable and non-editable, thus, making up a store of frozen, static text. This makes retrieval of information difficult. Optical Character Recognition (OCR) technology is a technology that turns this hard to read text into machine readable information thus making it searchable.

**Facilities:** Support Convenient Document Editing.
OCR is a multi-purpose tool that makes the business processes more flexible towards change. OCR facilitates a smooth transfer of changes by ensuring that files that cannot be changed are transformed into a text that can be easily edited. OCR is necessary in converting the non-editable documents into document types, which will be easily edited.

**Mitigate Human Error**
Human operators are bound to make mistakes but on the flipside, OCR allows editing and searching documents which are otherwise uneditable as well as detecting mistakes or the misprinting of data on them. As a result, the probability of human error can be eliminated in advance by using the OCR technology.

**Cut down Time and Financial Cost.**
Many organisations are still still depending on paper-based records. OCR significantly reduces the duration and the cost of entering data by hand into computer systems. OCR produces easily editable copies of the printed materials or images by just scanning, thus saving resources.

**Conserve Physical Space**
Paper records even waste office space. Digitising paper records by use of OCR releases space capacity. Consequently, less paper use will result in fewer papers on the workspace and an optimised utilisation of the physical resources.

**6. Conclusion**
The suggested methodology presents a text classification mechanism which could be used to deploy in governmental organizations, and it is an automated system.
Experimental analysis of total-text Dataset shows that there is a record accuracy of 94.62% in text recognition in images. The findings corroborate that the deployed text recognition and classification solutions represent an appropriate approach of verifying a general institution because they can be used in variant fields including education, finance, healthcare, and human resources.

The system is specifically appropriate due to low resource usage and lack of need in the external hardware.

**7. Future work**
This research can be extended in several directions. Some possible avenues are:
- **Low-resource and multilingual support**
  It is acutely required to create powerful models that can operate with languages or scripts that have a lack of labelled dataset, i.e., French, Russian, Arabic, etc. It is possible to achieve this undertaking with the aid of cross-lingual transfer learning, synthetic data generation, and multilingual pre-training. More assessment needs to be provided to determine the results of performance and quantification of the error rates in the language.
- **Label evolution and automatic classification induction**
  Apply approaches that suggest new classes automatically, group or bifurcate existing ones, which are proposed by both clustering and users, so that the taxonomy may adapt to the changing collection of text images.
- **Continuous learning and domain adaptation**
  We seek to develop strong methodologies that allow the adaptation of a model to new areas, including education, finance, human resources, and so on, with scarce labeled data, through the support of advanced paradigms, like unsupervised domain adaptation, few-shots, and continual learning; we also were developing stringent regularization and replay processes to prevent disastrous forgetting when new classes are introduced.

**References**
1. Kim G, Lee H, Park S, Jeon J, Park J, Seo M, *et al.* Ocr-free document understanding transformer. In: European Conference on Computer Vision. 2022. p. 498-517.
2. Li Q, Li M, Zhang T, Wu X, Chen H, Yang Y, *et al.* A survey on text classification: From traditional to deep learning. ACM Trans Intell Syst Technol. 2022;13(2):1-41.
3. Huang Y, Lv T, Cui L, Lu Y, Wei F. Layoutlmv3: Pre-training for document AI with unified text and image masking. In: Proceedings of the 30th ACM International Conference on Multimedia. 2022. p. 4083-4091.
4. Ch'ng CK, Chan CS. Total-text: A comprehensive dataset for scene text detection and recognition. In: 2017 14th IAPR International Conference on Document Analysis and Recognition (ICDAR). 2017. p. 935-942.
5. Long S, He X, Yao C. Scene text detection and recognition: The deep learning era. Int J Comput Vis. 2021;129(1):161-184.
6. Baek Y, Lee B, Han D, Yun S, Lee H. Character region awareness for text detection. In: Proceedings of the IEEE/CVF Conference on Computer Vision and Pattern Recognition. 2019. p. 9365-9374.
7. Ye J, Chen Z, Liu J, Du B. TextFuseNet: Scene text detection with richer fused features. In: IJCAI. 2020. p. 516-522.
8. Zhang S-X, Zhu X, Yang C, Wang H, Yin X-C. Adaptive boundary proposal network for arbitrary shape text detection. In: Proceedings of the IEEE/CVF International Conference on Computer Vision. 2021. p. 1305-1314.
9. Zhao F, Yu J, Xing E, Song W, Xu X. Real-time scene text detection based on global level and word level features. arXiv Preprint. 2022;arXiv:2203.05251:1-10.
10. Su Y, Wang L, Huang J, Zhao M, Li X, Zhou Q, *et al.* TextDCT: Arbitrary-shaped text detection via discrete cosine transform mask. IEEE Trans Multimedia. 2022;25:5030-5042.
11. You Y, Lei Y, Zhang Z, Tong M. Arbitrary-shaped text detection with B-Spline curve network. Sensors. 2023;23(5):2418-2428.
12. Chen Z. (HTBNet) Arbitrary shape scene text detection with binarization of hyperbolic tangent and cross-entropy. Entropy. 2024;26(7):560-570.
13. Horiuchi T. Grayscale image segmentation using color space. IEICE Trans Inf Syst. 2006;89(3):1231-1237.
14. Wang X, Xie L, Dong C, Shan Y. Real-ESRGAN: Training real-world blind super-resolution with pure synthetic data. In: Proceedings of the IEEE/CVF International Conference on Computer Vision. 2021. p. 1905-1914.
15. Musa P, Al Rafi F, Lamsani M. A review: Contrast-limited adaptive histogram equalization (CLAHE) methods to help the application of face recognition. In: 2018 Third International Conference on Informatics and Computing (ICIC). 2018. p. 1-6.
16. Liao M, Zou Z, Wan Z, Yao C, Bai X. Real-time scene text detection with differentiable binarization and adaptive scale fusion. IEEE Trans Pattern Anal Mach Intell. 2022;45(1):919-931.
17. Lewis M, Liu Y, Goyal N, Ghazvininejad M, Mohamed A, Levy O, *et al.* BART: Denoising sequence-to-sequence pre-training for natural language generation, translation, and comprehension. arXiv Preprint. 2019;arXiv:1910.13461:1-30.
18. Thanki JD, Davda PD, Swaminarayan P. A review on OCR technology. JETIR. 2021;8(4):1-12.
19. Zhou X, Zhang Y, Liu F, Wang H, Zhang L, Chen J, *et al.* A survey on text classification and its applications. In: Web Intelligence. 2020. p. 205-216.